\documentclass[runningheads]{llncs}
\usepackage[T1]{fontenc}

\usepackage{graphicx}

\usepackage{lineno}
\usepackage{url}

\usepackage{amssymb} %
\usepackage{amsmath}
\usepackage{tabularx}
\usepackage{caption}
\usepackage{array} %
\newcolumntype{Y}{>{\centering\arraybackslash}X} %

\usepackage{tikz}
\usepackage{float}
\DeclareRobustCommand{\trianglediagonal}{%
  \mathord{%
    \tikz[scale=0.12,baseline=-0.6ex]{%
      \fill (0.3,1) -- (1,1) -- (1,0.3) -- cycle;
      \draw (0,1) -- (1,0);
    }%
  }%
}
\renewcommand{\triangle}{%
  \mathord{%
    \begin{tikzpicture}[scale=0.12,baseline=-0.6ex]
      \fill (0.3,1) -- (1,1) -- (1,0.3) -- cycle;
    \end{tikzpicture}%
  }%
}
\usepackage{tikz}

\let\kt\relax %

\DeclareRobustCommand{\kt}{%
  \mathchoice{\ktdots{1.00}}{\ktdots{0.90}}{\ktdots{0.75}}{\ktdots{0.60}}%
}

\newcommand{\ktdots}[1]{%
  \mathord{%
    \tikz[baseline=-0.55ex, x=0.55mm, y=0.55mm, scale=0.5]{
      \foreach \x/\y in {
        3/0,3/2,
        4/1,4/3,
        5/0,5/2,
        6/1
      }{
        \fill (\x+0.5,\y+0.5) circle[radius=0.22mm];
      }
    }%
  }%
}

\usepackage{eso-pic}
\begin{document}
\AddToShipoutPictureFG*{\AtPageLowerLeft{\put(\LenToUnit{\dimexpr(\paperwidth-\textwidth)/2\relax},\LenToUnit{0.45in}){\parbox{\textwidth}{\footnotesize\centering Preprint (version submitted before peer review). Published in: Recent Trends and Challenges in Information Systems and Technologies (WorldCIST 2026), LNNS vol.~2092, Springer, \url{https://doi.org/10.1007/978-3-032-32023-0_31}}}}}
\title{Efficient Pairwise Difference Learning: A Comparative Study of Encoding Strategies and Training Pair Selection}
\titlerunning{Efficient Pairwise Difference Learning}
\author{Bernard Georges\inst{1} \and Mohamed Karim Belaid\inst{2} \and Manuel Rodrigues\inst{1}}
\authorrunning{B. Georges et al.}
\institute{University of Minho, ALGORITMI CENTER LASI, Braga, Portugal\\ \email{pg53698@uminho.pt, manuel.rodrigues@algoritmi.uminho.pt} \and IDIADA Fahrzeugtechnik GmbH, Munich, Germany\\ \email{karim.belaid@idiada.com}}

\maketitle  

\begin{abstract}
Pairwise Difference Learning for Classification (PDC) enhances supervised learning in low-data regimes by training a binary similarity classifier on pairs of instances rather than individual samples, effectively inflating the training set from n to O(n²) examples. However, the PDC framework leaves critical implementation choices underspecified, particularly regarding input representation and pair selection strategies. This work systematically investigates these design decisions through extensive empirical evaluation. We compare four feature encoders—paired features, triplet with relative difference, absolute difference, and a novel triplet with absolute difference—alongside four pair selection strategies ranging from exhaustive to kernel-thinning-based schemes. Experiments across nearly 200 benchmark classification datasets using Decision Tree, Multilayer Perceptron, Gaussian Naïve Bayes, and Support Vector Classifier as base models reveal that the proposed triplet with absolute difference encoder ($\phi_{\mathrm{3dabs}}$) achieves the highest average F1 scores. More notably, restricting training to upper-triangle pairs reduces computational time by approximately 40\% with negligible accuracy loss. Kernel thinning-based selection provided no accuracy benefits while substantially increasing computational overhead. Overall, PDC configurations outperformed base models in 94\% of tested cases, though no single configuration dominated universally, suggesting dataset-specific optimization remains an open challenge.

\end{abstract}
\section{Introduction}
Many machine learning applications operate in regimes where labeled examples are limited or expensive to acquire. In such settings, simply increasing model complexity is often less effective than enriching the supervision signal extracted from the available data. Pairwise learning schemes offer one such avenue: instead of learning from isolated samples, they create additional supervision by forming pairs of instances and encoding their relative relationships. Recent work has shown that pairwise methods can improve accuracy and robustness in regression and classification, especially in low-data regimes
\cite{vitynskyi2019,wetzel2020twin,izonin2021predictive,izonin2021additive,tynes2021pairwise,xie2022automated,corbara2023same,izonin2024adaptation,belaid2024pairwise,belaid2024weighted,wetzel2023how}.

In this work, we focus on Pairwise Difference Learning for classification (PDC) \cite{belaid2024pairwise}. PDC solves a multiclass classification task by training a binary ``similar vs.\ dissimilar'' classifier on pairs of examples and then using its outputs to reconstruct class posteriors. Intuitively, instead of asking the model to predict the label of a single point, PDC asks how likely two points are to share the same class. By pairing the $n$ training points, the method inflates the number of effective training examples to $\mathcal{O}(n^2)$, exposing the learner to rich comparative information that is particularly useful when data are scarce.

Formally, let $\mathcal{D}=\{(x_i,y_i)\}_{i=1}^n$ with $x_i \in \mathbb{R}^d$ and $y_i \in \{1,\dots,K\}$. PDC first selects a set of index pairs
\[
\mathcal{P} \subseteq \{1,\dots,n\}^2,\quad i \neq j,
\]
and applies a feature map
\[
\phi : \mathbb{R}^d \times \mathbb{R}^d \to \mathbb{R}^m,\qquad
z_{i,j} = \phi(x_i,x_j),
\]
Which encodes joint information about the two inputs (for example, their concatenation and/or their differences). Each pair receives a binary label.
\[
s_{i,j} = \mathbf{1}\{y_i = y_j\},
\]
Indicating whether the two instances share the same class. A base classifier $g$ is trained on the paired dataset.
\[
\mathcal{D}_{\mathrm{pair}} = \{(z_{i,j}, s_{i,j}) : (i,j)\in\mathcal{P}\},
\]
Learning to distinguish similar from dissimilar pairs. At prediction time, a query point $x$ is compared to a set of training anchors $\{x_j\}$, producing scores
\[
\hat{s}_{j}(x) = g\bigl(\phi(x,x_j)\bigr),
\]
which estimate how likely $x$ is to share its class with anchor $x_j$. Aggregating these scores over anchors of each class $k$ yields an estimate of the posterior class probabilities, from which the final prediction is obtained by selecting the class with the highest estimated probability.

\paragraph{Research gap.}
While the PDC framework is conceptually simple, it leaves several important implementation choices open. In practice, different PDC variants \cite{corbara2023same,izonin2024adaptation,belaid2024pairwise} differ at least along two dimensions:
(i) the input representation $\phi(x,x')$ used for the similarity model $g$, and
(ii) the definition of the pair set $\mathcal{P}$, that is, which pairs are generated and used for training and prediction. The original PDC variant of \cite{belaid2024pairwise} adopts a specific encoding and an almost exhaustive triplet-based pairing scheme, which can be computationally expensive and may include many weakly informative pairs. To the best of our knowledge, there is no systematic comparison of alternative input formats and pairing strategies within a unified experimental protocol.

\paragraph{Contributions.}
We build on the PDC formulation of \cite{belaid2024pairwise} and systematically study how these implementation choices affect its behaviour in practice. Our contributions are:
\begin{enumerate}
    \item We compare three different input formats for the similarity model in PDC, varying how the two instances and their differences are represented in $\phi(x,x')$, and study their impact on predictive performance and computational cost.
    \item We compare four pairing strategies for PDC, ranging from exhaustive to more selective schemes, and analyze how the choice of $\mathcal{P}$ influences accuracy, training time, and effective dataset size.
    \item We conduct an empirical evaluation on a suite of benchmark classification datasets, providing practical guidance on which PDC design choices are most effective under different data and resource constraints.
\end{enumerate}

\section{Related Work}

Early work by \cite{vitynskyi2019} proposed a pairwise scheme in which each instance was represented by its features along with a pairing weight, encoding similarity between vectors through a Gaussian kernel and feature-wise differences. Independently, \cite{wetzel2022twinb} introduced the twin-network paradigm. This approach employs two parallel neural networks with shared weights to project instances into a common latent space, where a distance function quantifies the degree of similarity or dissimilarity between inputs.

Subsequent contributions focused on structured and additive modeling. \cite{izonin2021predictive,izonin2021additive} developed pairwise strategies for structured data by applying Radial Basis Function (RBF) transformations to create augmented points, which are then paired with the original instances. In parallel and independently, \cite{tynes2021pairwise} worked on both the theoretical and practical scope of pairwise learning in chemical data, proposing the method later adopted in PDC, where instances are paired using both their original features and their differences.

Further advances addressed automation and scalability. \cite{xie2022automated} demonstrated the feasibility of scaling pairwise models to large, complex datasets, marking a shift toward practical applicability. This direction was reinforced by \cite{corbara2023same}, who emphasized evaluating pairwise models under heterogeneous and distributionally shifted data, using a pairing strategy based solely on absolute differences.%
\cite{wetzel2023how} offered a critical perspective by analyzing the theoretical limitations of pairwise methods, informing more principled model design. Both approaches incorporated point selection algorithms, leveraging random subsampling of the majority class and k-nearest neighbors, respectively.

The most recent work has emphasized refinement and adaptation. \cite{izonin2024adaptation} proposed adaptive pairwise strategies that allow models to dynamically adjust to evolving input distributions, highlighting the growing importance of flexibility in modern applications. Complementing this, \cite{belaid2024pairwise,belaid2024weighted} introduced pairwise weighting mechanisms to balance data contributions during training, advancing fairness, generalization, and interpretability in pairwise learning frameworks.

Regarding point selection, \cite{dwivedi2024kernel} proposed a coreset-based approach that reduces the original dataset by iteratively partitioning instances into two balanced subsets using kernel computations, followed by selecting the optimal subset as the resulting dataset. 

\section{Method}
\subsection{Feature representation of a pair of datapoints}
A central design choice in PDC is how to represent a pair of inputs before passing it to the similarity model $g$.  
Given two datapoints $x_i, x_j \in \mathbb{R}^d$, an encoder
\[
\phi : \mathbb{R}^d \times \mathbb{R}^d \rightarrow \mathbb{R}^m
\]
Maps the pair $(x_i,x_j)$ to a feature vector $z_{i,j} = \phi(x_i,x_j)$, which is then used to predict whether the two instances belong to the same class.  
Different encoders trade off information content, symmetry, and dimensionality, and may therefore lead to different accuracy–cost profiles.

In the following, we consider four natural encoders that are all compatible with the same PDC pipeline (same classifier $g$ and same aggregation rule). They differ only in how the pair $(x_i,x_j)$ is represented.

\paragraph{Paired features ($\phi_{\mathrm{2d}}$).}
The simplest encoder keeps only the original features of the two datapoints, without any explicit difference term, an approach used in several pairwise classification schemes \cite{izonin2021additive,izonin2021predictive,izonin2024adaptation}:
\begin{equation}
\phi_{\mathrm{2d}}(x_i,x_j) = \bigl(x_i,\, x_j\bigr) \in \mathbb{R}^{2d}.
\label{eq:phi2d}
\end{equation}
Here, the model is expected to infer useful similarity or difference patterns internally from the concatenated inputs.  
The dimensionality is $2d$, and the encoder exposes the full original feature vectors to the classifier without imposing any particular structure on their relationship.

\paragraph{Triplet with relative difference ($\phi_{\mathrm{3drel}}$).}
A common PDC-style representation \cite{tynes2021pairwise,belaid2024pairwise} extends this idea by explicitly adding the signed, component-wise difference between the two inputs:
\begin{equation}
\phi_{\mathrm{3drel}}(x_i,x_j) = \bigl(x_i,\, x_j,\, x_i - x_j\bigr) \in \mathbb{R}^{3d}.
\label{eq:phi3drel}
\end{equation}
This encoder concatenates both inputs and their relative difference.  
The representation is directional: swapping $i$ and $j$ changes the sign of the last $d$ components.  
The additional block provides the classifier with explicit information about how the two instances differ along each feature dimension.

\paragraph{Absolute difference ($\phi_{\mathrm{abs}}$).}
A more compact alternative, inspired by SAME-style representations \cite{corbara2023same}, discards the original features and keeps only the component-wise absolute difference:
\begin{equation}
\phi_{\mathrm{abs}}(x_i,x_j) = \bigl|x_i - x_j\bigr| \in \mathbb{R}^d,
\label{eq:phiabs}
\end{equation}
where subtraction and absolute value are applied element-wise.  
This encoder is invariant to the order of the pair, i.e., $\phi_{\mathrm{abs}}(x_i,x_j)=\phi_{\mathrm{abs}}(x_j,x_i)$, so symmetric pairs do not need to be generated.  
It has the lowest dimensionality among the encoders considered here, which may reduce computational cost, but it also removes information about the absolute position of $x_i$ and $x_j$ in feature space.

\paragraph{Triplet with absolute difference ($\phi_{\mathrm{3dabs}}$).}
Finally, we propose a non-directional triplet representation that combines the full original features with an order-invariant difference term:
\begin{equation}
\phi_{\mathrm{3dabs}}(x_i,x_j) = \bigl(x_i,\, x_j,\, |x_i - x_j|\bigr) \in \mathbb{R}^{3d}.
\label{eq:phi3dabs}
\end{equation}
Compared to $\phi_{\mathrm{3drel}}$, this encoder removes the sign of the difference while retaining both original feature vectors.  
It therefore preserves essentially the same information, but it no longer encodes which of the two points is “larger” along a given dimension.  
This can simplify the learning task for $g$ by making the representation less sensitive to the ordering of the pair, while still providing a rich description of both absolute and relative structure.

\paragraph{Illustrative example.}
To make the four encoders concrete, consider two Titanic passengers described by two numeric features:  
\emph{Age} (in years) and \emph{Fare} (ticket price in US dollars).  
Let
\[
x_i = (22,\, 8), \qquad x_j = (38,\, 72),
\]
so that
\[
x_i - x_j = (-16,\, -64), \qquad |x_i - x_j| = (16,\, 64).
\]
Table~\ref{tab:phi_example} shows the resulting representations.  
For the same underlying pair $(x_i,x_j)$, $\phi_{\mathrm{2d}}$ keeps only who the passengers are, $\phi_{\mathrm{abs}}$ keeps only how much they differ, $\phi_{\mathrm{3drel}}$ augments the concatenation with signed differences, and $\phi_{\mathrm{3dabs}}$ combines the concatenation with absolute differences in a single feature vector.

\begin{table}
\centering
\caption{Example of the four pair encoders on two Titanic passengers with features Age (years) and Fare (US\$).}
\label{tab:phi_example}
\begin{tabular}{|l|l|}
\hline
Encoder & Output vector $\phi(x_i,x_j)$ \\
\hline
$\phi_{\mathrm{2d}}(x_i,x_j)$      & $(22,\; 8,\; 38,\; 72)$ \\
$\phi_{\mathrm{3drel}}(x_i,x_j)$   & $(22,\; 8,\; 38,\; 72,\; -16,\; -64)$ \\
$\phi_{\mathrm{abs}}(x_i,x_j)$     & $(16,\; 64)$ \\
$\phi_{\mathrm{3dabs}}(x_i,x_j)$   & $(22,\; 8,\; 38,\; 72,\; 16,\; 64)$ \\
\hline
\end{tabular}
\end{table}

\subsection{Input pair selection}
Beyond the feature encoder $\phi$, PDC also depends on how many and which pairs are actually generated during training.  
Given a training set
\[
D = \{x_1,\dots,x_n\},
\]
The naïve construction forms all ordered pairs
\[
D \times D = \{(x_i,x_j) : 1 \le i,j \le n\},
\]
Which can be visualized as an $n \times n$ matrix of pairs.  
This “full grid’’ includes self-pairs $(x_i,x_i)$ on the diagonal and symmetric duplicates $(x_i,x_j)$ and $(x_j,x_i)$ off the diagonal.  
While this maximizes the number of training examples ($n^2$ pairs), it also introduces redundancy and increases computational cost.

In this work, we model pair selection as an operator.
\[
\delta : \{1,\dots,n\}^2 \to \{0,1\},
\]
which indicates whether a pair $(i,j)$ is kept ($\delta(i,j)=1$) or discarded ($\delta(i,j)=0$).  
We consider four patterns, which can be visualized as different “shadings’’ of the $n \times n$ pair matrix.

\paragraph{All pairs ($\boldsymbol{\delta_{\blacksquare}}$).}
The baseline pattern keeps all ordered pairs, including both symmetric counterparts and the diagonal:
\[
\mathcal{P}_{\blacksquare}
= \{(i,j) : 1 \le i,j \le n\},
\qquad
|\mathcal{P}_{\blacksquare}| = n^2.
\]
This setting maximizes the training set size but also maximizes redundancy between $(i,j)$ and $(j,i)$, especially when $\phi$ is symmetric (e.g., absolute-difference encoders).

\paragraph{Unordered pairs ($\boldsymbol{\delta_{\trianglediagonal}}$).}
To remove symmetric duplicates while keeping self-pairs, we restrict to the upper triangle (including the diagonal):
\[
\mathcal{P}_{\trianglediagonal}
= \{(i,j) : 1 \le i \le j \le n\},
\qquad
\bigl|\mathcal{P}_{\trianglediagonal}\bigr| = \frac{n(n+1)}{2}.
\]
Each unordered pair $\{i,j\}$ appears exactly once, and diagonal elements $(i,i)$ are retained.  
This preserves self-pairs, which can be interpreted as “perfect matches’’ and may serve as a useful reference for the similarity model.
\paragraph{non symmetric unordered pairs ($\boldsymbol{\delta_{\triangle}}$).}
A more aggressive reduction removes both symmetric duplicates and self-pairs by keeping only the strict upper triangle:
\[
\mathcal{P}_{\triangle}
= \{(i,j) : 1 \le i < j \le n\},
\qquad
\bigl|\mathcal{P}_{\triangle}\bigr| = \frac{n(n-1)}{2}.
\]
This pattern eliminates self-pairs and uses each unordered pair exactly once.  
It allows us to test whether self-pairs are actually beneficial or merely inflate the training set with easy, potentially uninformative examples.

\paragraph{Kernel thinning ($\boldsymbol{\delta_{\kt}}$).}
We additionally consider a data–dependent selection pattern called kernel thinning \cite{dwivedi2024kernel}, in contrast to $\delta_{\blacksquare}$, $\delta_{\trianglediagonal}$ and $\delta_{\triangle}$, which keep or discard pairs solely based on their position in the $n \times n$ grid. Kernel thinning selects a subset of points that is well spread with respect to a chosen kernel, with the goal of producing a smaller training set that better represents the underlying distribution, thereby reducing imbalance and mitigating overfitting. Formally, it takes the original training set $D$ of size $n$ and returns a thinned subset $D^{(m)}$ of size approximately $h(n) = n / 2^m$, where $m \in \mathbb{N}$ is a splitting parameter: each increment in $m$ doubles the number of thinning steps and halves the dataset size.

In our notation, $\delta_{\kt}$ denotes the mask induced by applying kernel thinning either before or after pair construction, and $\bigl|\mathcal{P}_{\text{kt}}\bigr|$ corresponds to the number of pairs that the similarity model receives.  
If thinning is applied \emph{before} pairing, we first obtain $D^{(m)}$ of size $h(n) = n/2^m$ data points and then build unordered pairs using $\delta_{\trianglediagonal}$. The resulting number of pairs is
\[
\bigl|\mathcal{P}_{\kt}^{\mathrm{pre}}\bigr|
= \frac{h(n) \cdot (h(n)+1)}{2}
= \frac{n\,(n+2^m)}{2^{2m+1}}.
\]
If thinning is applied \emph{after} pairing, we first construct $\mathcal{P}_{\trianglediagonal}$ of size $n(n+1)/2$ and then thin the set of pairs by the same factor $2^m$, yielding.
\[
\bigl|\mathcal{P}_{\kt}^{\mathrm{post}}\bigr|
= h \bigl( \bigl|\mathcal{P}_{\trianglediagonal}\bigr| \bigr)
= \frac{n(n+1)}{2^{m+1}}.
\]
In practice, we restrict $m$ to small values (here $m\in\{1,2\}$) to avoid excessively small training sets, and we primarily use pre-pairing thinning to better respect the i.i.d.\ assumption on the inputs. This data–adaptive mask $\delta_{\kt}$ therefore provides a fourth, complementary way to control the effective number of training pairs seen by PDC, trading off computational cost against distributional coverage.

\paragraph{Illustrative example.}
To visualize the effect of the different selection patterns, consider a toy dataset of four datapoints represented by the four card suits,
\[
D = (x_1,x_2,x_3,x_4) = (\spadesuit,\heartsuit,\diamondsuit,\clubsuit).
\]
Table~\ref{tab:delta_example} shows how $D \times D$ is transformed by the masks
$\delta_{\blacksquare}$, $\delta_{\trianglediagonal}$, $\delta_{\triangle}$, and $\delta_{\kt}$ (with $m=2$).  
The first three patterns are purely index-based: they keep all pairs, only the upper triangle with diagonal, or only the strict upper triangle, respectively.  
The kernel-thinning pattern $\delta_{\kt}$ instead keeps half of the pairs (here illustrated by an arbitrary subset of cells).

\begin{table}
\centering
\caption{Toy example of pair-selection patterns on $D=(\spadesuit,\heartsuit,\diamondsuit,\clubsuit)$. Each matrix entry denotes the ordered pair $(x_i,x_j)$; empty cells indicate discarded pairs. For $\delta_{\kt}$ with $m=1$, an arbitrary half of the pairs are kept.}
\label{tab:delta_example}
\begin{tabularx}{\textwidth}{cYYYY}
 $D$ 
 & $\delta_{\blacksquare}(D)$ 
 & $\delta_{\trianglediagonal}(D)$ 
 & $\delta_{\triangle}(D)$ 
 & $\delta_{\kt}^{post}(D)$, $m=1$
 \\
\hline
$\begin{bmatrix}
\spadesuit\\[2pt]
\heartsuit\\[2pt]
\diamondsuit\\[2pt]
\clubsuit
\end{bmatrix}$
&
$\begin{bmatrix}
\spadesuit\spadesuit & \spadesuit\heartsuit & \spadesuit\diamondsuit & \spadesuit\clubsuit \\
\heartsuit\spadesuit & \heartsuit\heartsuit & \heartsuit\diamondsuit & \heartsuit\clubsuit \\
\diamondsuit\spadesuit & \diamondsuit\heartsuit & \diamondsuit\diamondsuit & \diamondsuit\clubsuit \\
\clubsuit\spadesuit & \clubsuit\heartsuit & \clubsuit\diamondsuit & \clubsuit\clubsuit
\end{bmatrix}$
&
$\begin{bmatrix}
\spadesuit\spadesuit & \spadesuit\heartsuit   & \spadesuit\diamondsuit & \spadesuit\clubsuit \\
                     & \heartsuit\heartsuit   & \heartsuit\diamondsuit & \heartsuit\clubsuit \\
                     &                         & \diamondsuit\diamondsuit & \diamondsuit\clubsuit \\
                     &                         &                         & \clubsuit\clubsuit
\end{bmatrix}$
&
$\begin{bmatrix}
\phantom{\spadesuit\spadesuit} & \spadesuit\heartsuit   & \spadesuit\diamondsuit & \spadesuit\clubsuit \\
                     &                         & \heartsuit\diamondsuit & \heartsuit\clubsuit \\
                     &                         &                         & \diamondsuit\clubsuit \\
                     &                         &                         &
\end{bmatrix}$
 &
$\begin{bmatrix}
\spadesuit\spadesuit & \spadesuit\heartsuit &                         &                      \\
                     &                     &                          & \heartsuit\clubsuit  \\
                     &                     & \diamondsuit\diamondsuit &                     \\
                     &                     &                          & \clubsuit\clubsuit
\end{bmatrix}$
\end{tabularx}
\end{table}

\section{Evaluation} %
\subsection{Benchmarking datasets}
The datasets used in these experiments are from the OpenML repository, \cite{OpenML2021}. These datasets were selected and used by \cite{belaid2024pairwise}, and in order to ensure consistency and continuity across all experimental runs, the PDC application contains a parquet file that contains the IDs of the used datasets. This list is then used to retrieve the datasets needed from the OpenML repository. Furthermore, in order to confirm the use of the same datasets across all analysis, in case of errors or other random events during execution, the meta-algorithm extracts key descriptors from the API and dataset, including the number of features, number of instances, class distribution and other 8 descriptor values that are then cross-verified to ensure that the identical datasets will be used across all comparisons and analyses done with the results. 

PDC also performs some basic pre-processing on the datasets utilising the methods available in the \cite{scikit-learn} library. Namely, it uses \textit{OrdinalEncoder, OneHotEncoder} and \textit{StandardScaler}. After which, it drops columns containing null values. In the wake of the pre-processing, the data is then put through the cross-validation method that divides it into 5 folds. These steps were not altered, meaning that a faithful reproduction can be obtained through the algorithm presented in the original repository.

\subsection{Base ML models}
PDC is a meta-algorithm, meaning that it works on top of multiple ML models. These models have different paradigms, meaning different requirements and strategies. Due to this, 4 paradigms were chosen, each represented by a model. Decision Tree is our baseline model, meaning it will be used to test all initial approaches; the best results will then also be used with the other 3 models.

\begin{itemize}
    \item \textbf{Decision Tree (DT)} - Its interpretability, simplicity, and ability to handle both categorical and numerical data make it one of the most widely used models.
    \item \textbf{Multilayer Perceptron Classifier (MLP)} - MLP is a neural network. These models are capable of approximating highly non-linear decision boundaries
    \item \textbf{Gaussian Naïve Bayes Classifier (GNB)} - A probabilistic model, meaning it computes the likelihood of each possible class and assigns the instance to the class with the highest probability.
    \item \textbf{Support Vector Classifier (SVC)} -  based on the principles of Support Vector Machines (SVMs), it aims to find an optimal separating hyperplane in the feature space that maximises the margin between classes. 
\end{itemize}

\subsection{Metric}
This study focused primarily on two metrics: accuracy and computational time. For accuracy, the selected metric was the $F_1$ score as it provides a balanced and interpretable measure of model performance and is particularly suited for evaluating imbalanced datasets, such as those generated by the PDL algorithm. The analysis further extends beyond individual dataset-specific $F_1$ scores to include the mean $F_1$ score, $\bar{F_1}$, computed across all datasets within a given experiment. As for computational performance, it is primarily assessed through the total execution time of each run, including training and testing time. To further validate the results, following the identification of the most effective improvements on these metrics, a t-test and Bayesian posterior coefficients are conducted to determine the statistical significance and validity of the results with 95\% significance.

\subsection{Results \& Discussion}
\begin{table}[H]
\centering
\begin{tabular}{|l|c|c|c|c|}
\hline
& & $\phi_{\mathrm{3drel}}$ & $\phi_{\mathrm{abs}}$ & $\phi_{\mathrm{3dabs}}$ \\
\hline
All Pairs ($\boldsymbol{\delta_{\blacksquare}}$)& $F_1$ & 0.804 $\pm$ 0.151 & 0.797 $\pm$ 0.154 & 0.810 $\pm$ 0.150 \\
\cline{2-5}
& Time [min] & 81 & 68 & 81 \\
\hline
UT w/ diagonal ($\boldsymbol{\delta_{\trianglediagonal}}$)& $F_1$ & 0.806 $\pm$ 0.148 & 0.796 $\pm$ 0.155 & 0.810 $\pm$ 0.148 \\
\cline{2-5}
& Time [min] & 43 & 50 & 44 \\
\hline
UT w/o diagonal ($\boldsymbol{\delta_{\triangle}}$)& $F_1$ & 0.806 $\pm$ 0.148 & N/A & 0.809 $\pm$ 0.147 \\
\cline{2-5}
& Time [min] & 44 & N/A & 48 \\
\hline
\end{tabular}
\caption{$F_1$ score Comparison of different input combinations}
\label{tab:ut-results}
\end{table} 

As reported in Table~\ref{tab:ut-results}, the $\phi_{\mathrm{3dabs}}$ method achieved the highest average $F_1$ scores. This was achieved both with $\boldsymbol{\delta_{\blacksquare}}$ and $\boldsymbol{\delta_{\trianglediagonal}}$ yielding 0.810, meaning improvements of 0.006 and 0.004 over the baseline, $\phi_{\mathrm{3drel}}$, respectively. 

\begin{figure}[H]
    \centering
    \includegraphics[width=0.75\linewidth]{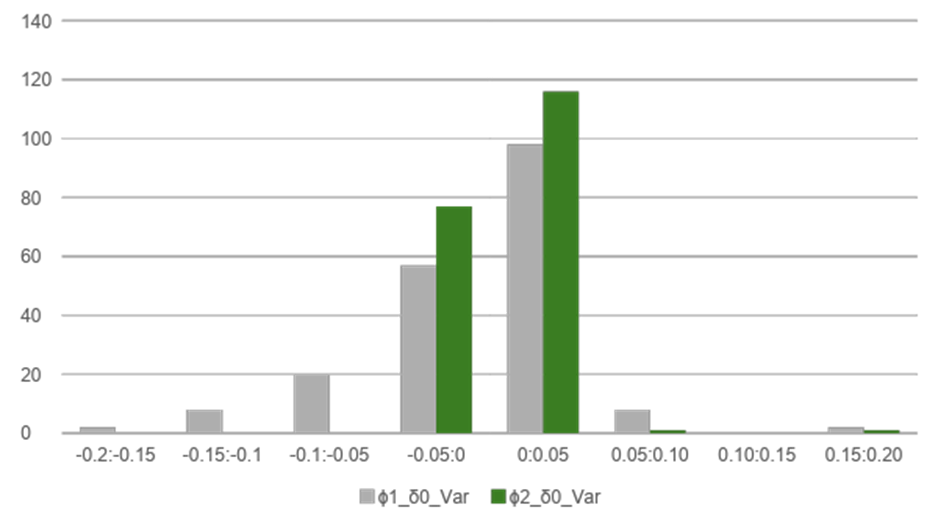}
    \caption{Histogram of the $F_1$ score difference of between a method and $\phi_{\mathrm{3drel}}$ \protect\footnotemark[1]} 
    \label{fig:histogram_difference}
\end{figure} 

\footnotetext[1]{The notation for this figure is based on the original work with: $\phi_1 =  \phi_{\mathrm{abs}}$, $\phi_2 = \phi_{\mathrm{3dabs}}$ and $\delta_0 = \boldsymbol{\delta_{\blacksquare}}$}

Nonetheless, when closely examined and as presented in figure~\ref{fig:histogram_difference}, $\phi_{\mathrm{abs}}$ presents higher amounts of big improvements, improvements greater than 0.05, while also having more deterioration, negative differences. These differentials demonstrate that even though $\phi_{\mathrm{3drel}}$ presents higher averages than $\phi_{\mathrm{abs}}$, the later still presents good functionalities in specific datasets. Furthermore, when examined closely and subjected to a paired statistical test such as Student $t$-test, no statistically significant effects were detected meaning the improvements may have been too inconsistent. 

The most substantial benefit of these representations lies instead within the $\boldsymbol{\delta_{\trianglediagonal}}$ method, which consistently presented the best or a slightly lower $F_1$ score with an average of 40\% decrease on the time taken to accomplish the experiments.

\begin{table}[h!]
\centering
\begin{tabular}{|c|c|c|c|c|}
\hline
 & Baseline & Before Pairing (m=1) & After Pairing (m=1) & After Pairing (m=2) \\
\hline
Results (80 DS) & 0.838 & 0.811 & 0.838 & 0.825 \\
\hline
Time [min]  & 7 & 7 & 209 & 147 \\
\hline
\end{tabular}
\caption{Comparison of $F_1$ and run time across Different Kernel Thinning Settings for 80 Datasets per experiment}
\end{table}

For the point selection, the results did not present any improvements, with the best case when selection is done after pairing. Nonetheless, in this case, the computation complexity presents a major increase, with an increase of 290\% or 207 minutes, and no improvement in the accuracy. 

With these results in mind, the best improvements and chosen methods to continue experimentation on the remaining 3 models were $\phi_{\mathrm{3dabs}}$ combined with $\boldsymbol{\delta_{\blacksquare}}$ and $\boldsymbol{\delta_{\trianglediagonal}}$, and $\phi_{\mathrm{2d}}$.

\begin{table}[H]
\begin{tabular}{|l|c|c|c|c|c|}
\hline
 Model (num of datasets)& No PDL & $ \phi_{\mathrm{3drel}} \circ \boldsymbol{\delta_{\blacksquare}}$ & $\phi_{\mathrm{3dabs}} \circ \boldsymbol{\delta_{\blacksquare}}$ & $ \phi_{\mathrm{3dabs}} \circ \boldsymbol{\delta_{\trianglediagonal}}$ & $ \phi_{\mathrm{2d}} \circ \boldsymbol{\delta_{\blacksquare}}$\\
\hline
 MLP (187 DS) & 0.759 $\pm$ 0.168 & 0.794 $\pm$ 0.168 & 0.810 $\pm$ 0.159 & 0.803 $\pm$ 0.162 & 0.794 $\pm$ 0.169\\
\hline
 GaussianNB (195 DS)  & 0.727 $\pm$ 0.179 & 0.603 $\pm$ 0.230 & 0.691 $\pm$ 0.212 & 0.684 $\pm$ 0.215 & 0.327 $\pm$ 0.088\\
\hline
 SVC (70 DS) & 0.869 $\pm$ 0.152 & 0.871 $\pm$ 0.157 & 0.880 $\pm$ 0.141 & 0.880 $\pm$ 0.144 & N/A \\
\hline
 Decision Tree (195 DS) & 0.782 $\pm$ 0.152 & 0.804 $\pm$ 0.151 & 0.810 $\pm$ 0.150 & 0.810 $\pm$ 0.148 & N/A\\
 \hline
 \end{tabular}
\caption{$F_1$ score of every experiment done in the different models}
\label{tab:comparison_othermodels}
\end{table}

MLP presented the best $F_1$ improvement with 0.016 for $\phi_{\mathrm{3dabs}} \circ \boldsymbol{\delta_{\blacksquare}}$ this was closely followed by $\phi_{\mathrm{3dabs}} \circ \boldsymbol{\delta_{\trianglediagonal}}$ with 0.803. These results presented the general trend of all the results with $\phi_{\mathrm{3dabs}} \circ \boldsymbol{\delta_{\blacksquare}}$ having the highest score, closely followed by 
$\phi_{\mathrm{3dabs}} \circ \boldsymbol{\delta_{\trianglediagonal}}$ with either the same $F_1$ or a slight decrease. This came with great improvements in time and complexity. In the case of MLP, there was a 54\% decrease in time. These results, similarly to what happened to DT, did not present significance with the $t$-test. Nevertheless, they did present a Bayesian probability higher than 75\% of being better than $ \phi_{\mathrm{3drel}} \circ \boldsymbol{\delta_{\blacksquare}}$, which is significant.

As for GNB, the result showed the highest increase with an increase of 0.088 in $F_1$ between $ \phi_{\mathrm{3drel}} \circ \boldsymbol{\delta_{\blacksquare}}$ and $\phi_{\mathrm{3dabs}} \circ \boldsymbol{\delta_{\blacksquare}}$. Nonetheless, this improvement is not enough to beat the base model without PDC, this being the only case where PDC presented a deterioration when compared to the model. 

SVC maintains the trend with the highest $F_1$ result. This model, however, also showed the highest amount of prediction errors, with 10 datasets not being able to complete their predictions due to the model not predicting any instance as the class. 

\begin{figure}[H]
    \centering
    \begin{minipage}{0.49\textwidth}
        \centering
        \includegraphics[width=\linewidth]{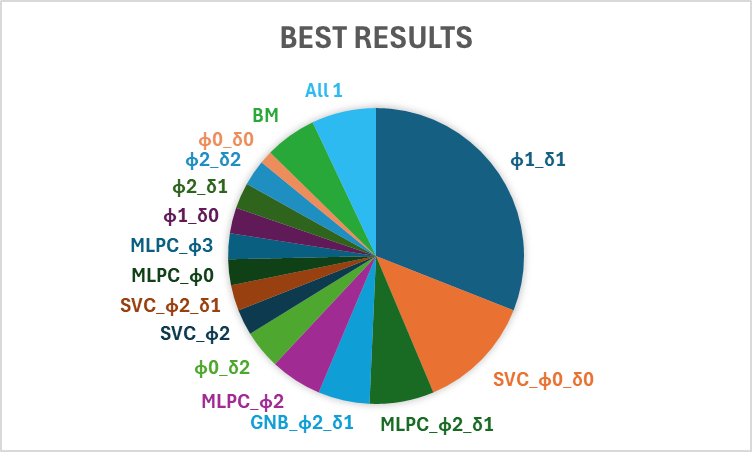}
        \caption{Pie Chart of All Methods with Best Results per Dataset\protect\footnotemark[2]}
        \label{fig:pie_detailed}
    \end{minipage}
    \hfill
    \begin{minipage}{0.49\textwidth}
        \centering
        \includegraphics[width=\linewidth]{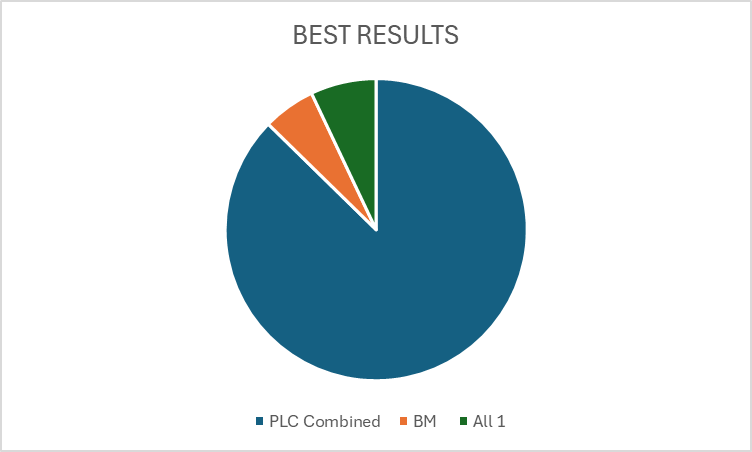}
        \caption{Pie Chart of the PDC Results vs Base Model Results per Dataset\protect\footnotemark[2]}
        \label{fig:pie_general}
    \end{minipage}
\end{figure}

\footnotetext[2]{The notation for this figure is based on the original work with: $\phi_0 =  \phi_{\mathrm{3drel}}$ $\phi_1 =  \phi_{\mathrm{abs}}$, $\phi_2 = \phi_{\mathrm{3dabs}}$, $\delta_0 = \boldsymbol{\delta_{\blacksquare}}$, $\delta_1 = \boldsymbol{\delta_{\trianglediagonal}}$ and $\delta_2 = \boldsymbol{\delta_{\triangle}}$}

For a final analysis, it was comprised all the $F_1$ scores across all the common datasets done through the experiment. The chart shows the amount of datasets in which each method presented the best result with a precision of 5 decimal points. As seen in Figure~\ref{fig:pie_general} in its entirety PDC generally presented improvements in the $F_1$ score, with only 4 out of 70 cases reporting a better score in the base model. However, as presented in Figure~\ref{fig:pie_detailed}, no single method consistently presents the best score, with the highest amount being presented by $\phi_{\mathrm{abs}} \circ \boldsymbol{\delta_{\trianglediagonal}}$, 22 datasets which is 30\% of the total datasets. This example further demonstrates the inconsistency of such methods, which achieve great improvements while also having a low averages. 

\section{Conclusion}

A closer inspection of the prediction evaluation revealed that a significant improvement is possible on a per-dataset basis, where the PDL, when evaluated collectively in its different configurations, was able to reach the best $F_1$ score in most tested cases. Nevertheless, no single configuration can be distinguished as each presents some datasets in which it outperforms its peers. To pursue the idea of an optimiser that selects the most adequate configuration for a given field of application or dataset class is once again beyond the scope of this study.

In spite of the absence of statistical significance, the obtained results provided good insights into the working of the PDC algorithm, particularly in the design and evaluation of its training phase. During the study, some of the proposed modifications presented promising results, particularly with the pairing function $\phi_{\mathrm{3dabs}}$ demonstrating strong performance in three cases, namely when the meta-algorithm used either the DT, MLP, or SVC as its underlying base models. Moreover, the investigation into the training scope has also revealed an interesting trade-off. As training in the upper triangle, $\delta_{\trianglediagonal}$, has shown a statistically non-significant decrease in accuracy when compared to its full matrix counterpart, $\delta_{\blacksquare}$, it significantly decreased the complexity and time spent in training.

On the other hand, the research experimented with the use of point selection during the training phase, which was subsequently discarded as a valid enhancement feature. In fact, the point selection method resulted in a great increase in time with no gratification presented in return for its drawback. Beyond this, the results corroborated the finding mentioned in \cite{belaid2024pairwise} regarding the ineffectiveness of the GNB classifier as a base model for the PDC meta-algorithm. The collected results presented with 95\% certainty a significant decrease in $F_1$ scores when compared to other base models. 

\bibliographystyle{splncs04}
\bibliography{bibliography}

\begin{thebibliography}{10}
\providecommand{\url}[1]{\texttt{#1}}
\providecommand{\urlprefix}{URL }
\providecommand{\doi}[1]{https://doi.org/#1}

\bibitem{belaid2024pairwise}
Belaid, M.K., Rabus, M., H{\"u}llermeier, E.: Pairwise difference learning for
  classification. In: International Conference on Discovery Science. pp.
  284--299. Springer (2024)

\bibitem{belaid2024weighted}
Belaid, M.K., Wibiral, T., Rabus, M., H{\"u}llermeier, E.: Weighted {Pairwise
  Difference Learning}. In: proc.\ of LWDA'24. CEUR-WS.org, Würzburg (2024)

\bibitem{OpenML2021}
Bischl, B., Casalicchio, G., Feurer, M., Gijsbers, P., Hutter, F., Lang, M.,
  Mantovani, R.G., van Rijn, J.N., Vanschoren, J.: Openml: A benchmarking layer
  on top of openml to quickly create, download, and share systematic
  benchmarks. NeurIPS  (2021),
  \url{https://datasets-benchmarks-proceedings.neurips.cc/paper_files/paper/2021/hash/c7e1249ffc03eb9ded908c236bd1996d-Abstract-round2.html}

\bibitem{corbara2023same}
Corbara, S., Moreo, A., Sebastiani, F.: Same or different? diff-vectors for
  authorship analysis. ACM Transactions on Knowledge Discovery from Data
  \textbf{18}(1),  1--36 (2023)

\bibitem{dwivedi2024kernel}
Dwivedi, R., Mackey, L.: Kernel thinning. Journal of Machine Learning Research
  \textbf{25}(152),  1--77 (2024)

\bibitem{izonin2021predictive}
Izonin, I., Tkachenko, R., Dronyuk, I., Tkachenko, P., Gregus, M., Rashkevych,
  M.: Predictive modeling based on small data in clinical medicine: Rbf-based
  additive input-doubling method. Math. Biosci. Eng.  (2021)

\bibitem{izonin2024adaptation}
Izonin, I., Tkachenko, R., Havryliuk, M., Gregus, M., Yendyk, P., Tolstyak, Y.:
  An adaptation of the input doubling method for solving classification tasks
  in case of small data processing. Procedia Computer Science  (2024)

\bibitem{izonin2021additive}
Izonin, I., Tkachenko, R., Shakhovska, N., Lotoshynska, N.: The additive
  input-doubling method based on the svr with nonlinear kernels. Symmetry
  (2021)

\bibitem{scikit-learn}
Pedregosa, F., Varoquaux, G., Gramfort, A., Michel, V., Thirion, B., Grisel,
  O., Blondel, M., Prettenhofer, P., Weiss, R., Dubourg, V., Vanderplas, J.,
  Passos, A., Cournapeau, D., Brucher, M., Perrot, M., Duchesnay, E.:
  Scikit-learn: Machine learning in {P}ython. Journal of Machine Learning
  Research  \textbf{12},  2825--2830 (2011)

\bibitem{tynes2021pairwise}
Tynes, M., Gao, W., Burrill, D.J., Batista, E.R., Perez, D., Yang, P., Lubbers,
  N.: Pairwise difference regression: a machine learning meta-algorithm for
  improved prediction and uncertainty quantification in chemical search.
  Journal of chemical information and modeling  \textbf{61}(8),  3846--3857
  (2021)

\bibitem{vitynskyi2019}
Vitynskyi, P., Tkachenko, R.: Neiropodibna struktura dlya zadach prohnozuvannya
  v umovakh korotkykh vyborok danykh. Sci. Bull. UNFU  (2019)

\bibitem{wetzel2023how}
Wetzel, S.J.: How to get the most out of {Twinned Regression} (Jan 2023)

\bibitem{wetzel2020twin}
Wetzel, S.J., Ryczko, K., Melko, R.G., Tamblyn, I.: Twin neural network
  regression. Applied AI Letters  \textbf{3}(4), ~e78 (2020)

\bibitem{wetzel2022twinb}
Wetzel, S.J., Ryczko, K., Melko, R.G., Tamblyn, I.: Twin neural network
  regression. Applied AI Letters  \textbf{3}(4), ~e78 (2022)

\bibitem{xie2022automated}
Xie, J., Kaiwei, C., Kong, L., Zhou, J., Qu, W.: Automated essay scoring via
  pairwise contrastive regression. In: proc.\ of the 29th COLING (2022)

\end{thebibliography}

\end{document}